\documentclass[letterpaper, 10 pt, conference]{ieeeconf}  

\IEEEoverridecommandlockouts                              

\usepackage{cite}
\usepackage{amsmath,amssymb,amsfonts}
\usepackage{algorithmic}
\usepackage{graphicx}
\usepackage{textcomp}
\usepackage{xcolor}

\usepackage{graphicx}
\usepackage{float}
\usepackage{subfig}
\usepackage{overpic}
\usepackage{hyperref}
\hypersetup{
    colorlinks=true,
    linkcolor=blue,
    filecolor=magenta,      
    urlcolor=cyan,
    pdftitle={Overleaf Example},
    pdfpagemode=FullScreen,
    }

\begin{document}
\title{\LARGE \bf
Development of a Simple and Novel Digital Twin Framework for Industrial Robots in Intelligent Robotics Manufacturing
}

\author{Tianyi Xiang$^{\&,1}$, Borui Li$^{\&,1}$, Xiaonan Pan$^{1}$, Quan Zhang$^{1,2,*}$
\thanks{The video for verification of the result can be accessed by:\href{https://youtu.be/f_BEMbMvFso?si=prVv36vperMNe9sU}{Video}}
\thanks{$^{\&}$These authors contributed equally to this work and should be considered co-first authors} 
\thanks{$^{1}$Authors are with School of Advanced Technology, Xi'an Jiaotong-Liverpool University, 111 Ren’ai Road, Suzhou, 215000.}%
\thanks{$^{2}$ Authors are with Department of Electrical and Electronic Engineering, University of Liverpool, Liverpool, L69 7ZX, UK.}
\thanks{$^{*}$ Email: Quan.Zhang@xjtlu.edu.cn}%
}

\maketitle

\begin{abstract}

This paper has proposed an easily replicable and novel approach for developing a Digital Twin (DT) system for industrial robots in intelligent manufacturing applications. Our framework enables effective communication via Robot Web Service (RWS), while a real-time simulation is implemented in Unity 3D and Web-based Platform without any other 3rd party tools. The framework can do real-time visualization and control of the entire work process, as well as implement real-time path planning based on algorithms executed in MATLAB. Results verify the high communication efficiency with a refresh rate of only $17 ms$. Furthermore, our developed web-based platform and Graphical User Interface (GUI) enable easy accessibility and user-friendliness in real-time control.

\end{abstract}

\section{Introduction}

In the era of Industry 4.0, manufacturing enterprises are moving towards intelligent manufacturing, to tackle, nowadays under great pressure to reduce costs and respond rapidly to the changing marketing environment, such as shorter product lifecycles, stochastic orders, customized individualized products, and the unplanned and disruptive manufacturing process, etc \cite{cimino2019review} \cite{lu2020digitalb} \cite{fuller2020digital}. For these challenges, great efforts have been devoted to developing cyber-physical production systems (CPPS) \cite{8477101}, with the most typical form of digital twin systems.

DT and the related technology have now been widely explored and applied to intelligent manufacturing systems, including industrial robotic systems, for real-time design, planning and optimization of the systems and their manufacturing processes, as well as real-time visualization and monitoring of the working processes 
\cite{10260401} \cite{9926434}\cite{martinez-gutierrez2021digital} \cite{s23010017}.

Despite much progress has been reached thus far, there are still many problems remaining unsolved with currently available technologies. Typically, there does not appear to have a relatively general approach suitable to a wide range of manufacturing applications, while the development of relevant DT is now still conducted catering for particular cases. Furthermore, the efficiency of currently available technologies, especially in the context of communication and transmission of a large amount of data, is still low in many studies \cite{OPCUalowrefreshrate} \cite{li2021automatic}. Whilst existing work, mostly, does not explicitly report relevant results, thus the efficiency issue of currently developed technologies remains questionable\cite{10260401}  \cite{9926434} \cite{martinez-gutierrez2021digital}\cite{s23010017} \cite{li2021automatic}\cite{zhu2022robot} \cite{wenna2022digital}. Hence, work for continuous improvements of manufacturing DT systems, in terms of generality of approaches, efficiency, user-friendly aspects, etc, are always necessary, which forms the subject matter of the present study.

The present study aims to explore approaches for manufacturing DT systems development, which are simple and user friendly, and also efficient and can be more generally applicable. Our study is based on a small-scale robot manufacturing station initially designed for grasping/handling tasks. This paper has developed a DT framework that enables remarkably effective real-time monitoring and optimization of the stacking process, including: (1) real-time visualization with a DT of the physical platform in Unity 3D and a Web App that can be accessed from anywhere regardless of distance; (2) effective data communication interfaces, developed by combining C\# and RWS – a combination adhering to Rest API and HTTP protocols \cite{robot}, with very simple mechanism, i.e. without any 3rd party frameworks such as Robot Operating System (ROS), no costly hardware devices like Programmable Logic Controller (PLC) are required; (3) enabling GUI in conjunction with MATLAB, for real-time process optimization solver (i.e. trajectory planning of robot arm). This renders our approach user-friendly as users can control the working station in real-time using a dynamic GUI with most widely used engineering tool, MATLAB, in a virtual environment.

\section{Related Work}

In recent years, the application of DT technology in intelligent manufacturing and digitalization of manufacturing systems become the trend in both industrial practice and scientific research, especially in robotic systems which are essential building blocks in intelligent manufacturing and, thus, key enablers of Industry 4.0 \cite{cimino2019review}. DT related technologies have now been applied for various aspects such as modelling and simulation, data management, connectivity and intelligence in intelligent manufacturing systems\cite{cimino2019review}.

Among the aforementioned aspects in DT, connectivity with proper communication approaches should be the most important since it directly affects the efficiency in operation of DT systems. Available methods include those establishing communication in DT based on OPC UA, ROS and PC-SDK, etc. Of the existing work developing and deploying DT in intelligent manufacturing systems including industrial robots, ROS has been a popular framework used for modelling and simulation, as well as establishing communication between the physical manufacturing systems and their virtual counterpart \cite{10260401}\cite{martinez-gutierrez2021digital}. It should, however, be noted that the use of ROS can introduce more challenges in developing DT systems due to the complexity in communication among nodes in ROS ecosystems, nor is ROS a tool that can be general to manufacturing applications since it is specifically developed for only robots. 

Studies of \cite{9926434} \cite{OPCUalowrefreshrate} apply OPC UA protocol for communication in their developed DT of industrial automatic manufacturing systems, OPC UA \cite{2023opc} is prevalent in industrial automation as it defines a universal communication protocol for control devices, e.g. PLCs, from different vendors. Nevertheless, deployment of OPC UA requires the control devices to be new, or additional hardware modules specifically supporting this protocol. Thus, it can add complexity and expense for system development, nor is it generally suitable for all manufacturing systems, especially those using old hardware controllers. Notably, study \cite{OPCUalowrefreshrate} that use OPC UA and ROS as communication protocol has mentioned the their best response time is $300ms$, while the subscription function is even 1 second response time. This reveals that the efficiency is very low, and the use of OPC UA adds complexity and increases system development costs.

PC-SDK, on the other hand, is a tool for secondary development on software for ABB robots DT \cite{PCSDK}. In our earlier work \cite{li2021automatic}, a DT system that relies on PC-SDK has been developed for the same workstation of this paper. The communication efficiency is very low, with the refreshing rate of the virtual model above 1 second in real-time monitoring without PLC. As the existing study does not appear to have resolved the issues in related approaches, further work is necessary which will be taken care of in this present paper.

DT and relevant technologies have been widely adopted in industrial robotics applications, such as Conveying \cite{10260401}, assembly \cite{9926434}, and collaborative painting \cite{s23010017}. Apart from the aspects of communication, effective approaches in modelling and simulation are also important since it is the most fundamental task creating a DT for any hardware system. Related existing studies apply either ROS \cite{martinez-gutierrez2021digital} or CoppeliaSim \cite{s23010017}, both are specifically devised for robots, for modelling of virtual robots. For simulation, path planning with virtual robots, considering collision avoidance \cite{zhu2022robot} and complicated 3D path planning \cite{wenna2022digital}, has been extensively studied and adopted to simulate the robot motion before implementation with physical robots, as well as to achieve real-time remote control of physical robots with DT. Nevertheless, the existing studies \cite{10260401} \cite{9926434} \cite{s23010017}  \cite{martinez-gutierrez2021digital} \cite{li2021automatic}\cite{zhu2022robot} \cite{wenna2022digital} almost all rely on specific tools and frameworks for modelling and simulation, which lacks generality in the context of approaches realizing DT. Again, the efficiency of the whole system, e,g, refreshing rate of the data driving the virtual models, is not reported. It is therefore, believed that developing an approach that can be relatively general to the majority of industrial automatic or intelligent manufacturing systems is always necessary, which will also be attempted in the present study. 

\begin{figure}
    \centering
    \includegraphics[width=\linewidth]{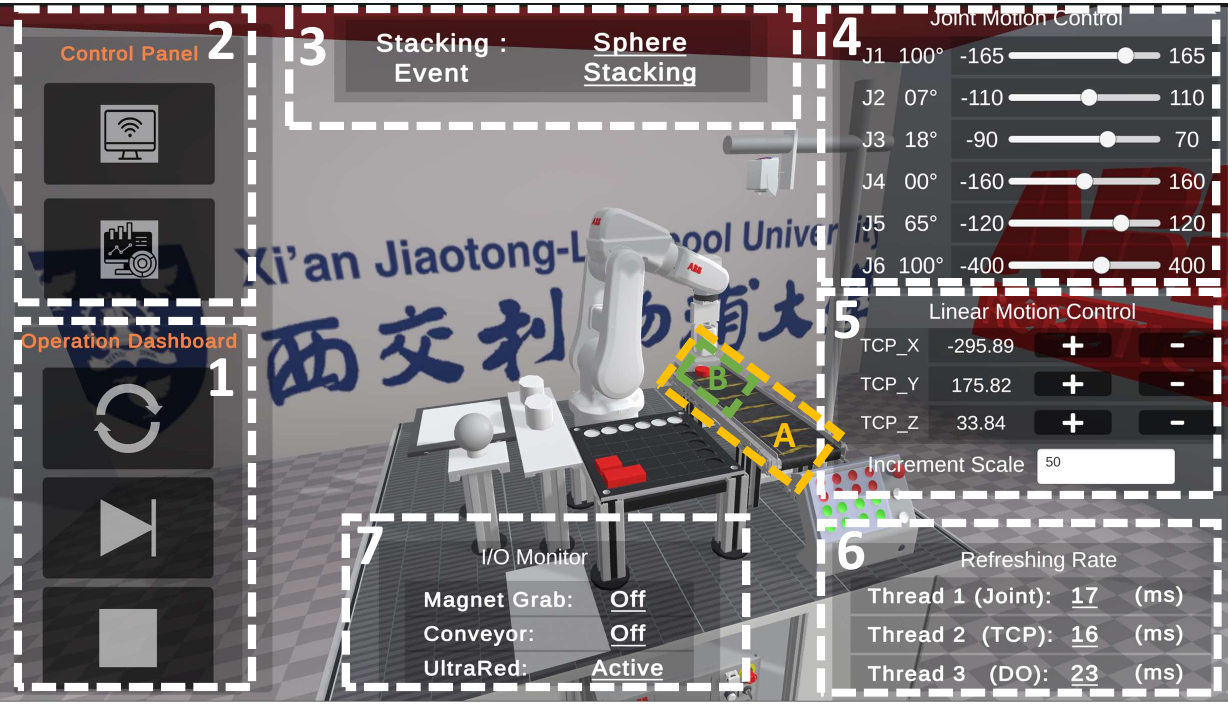}
    \caption{Digital Twin (Unity 3D) Working Screenshot and GUI Design (1st. Pointer Operation Panel; 2nd. IP Control Panel; 3rd. Event Monitoring; 4th. Joint Motion Control; 5th. Linear Motion Control; 6th. Refresh Rate Monitoring; 7th. I/O System Monitoring).}
    \label{GUI}
\end{figure}

\section{Framework}

Our developed DT framework of the stacking robot workstation, as shown in Fig. \ref{Framework}, include mainly three components: (1) Physical twin; and (2) Digital twin (Unity 3D); and (3) Web-Based App. Dual-directional communications and data transmission have been readily established among the aforementioned 3 components. Details of these components and their relationship inside our developed DT framework are now elaborated.

\subsection{The Robotic Manufacturing Station}

\begin{figure*}
    \centering
    \includegraphics[width = \linewidth]{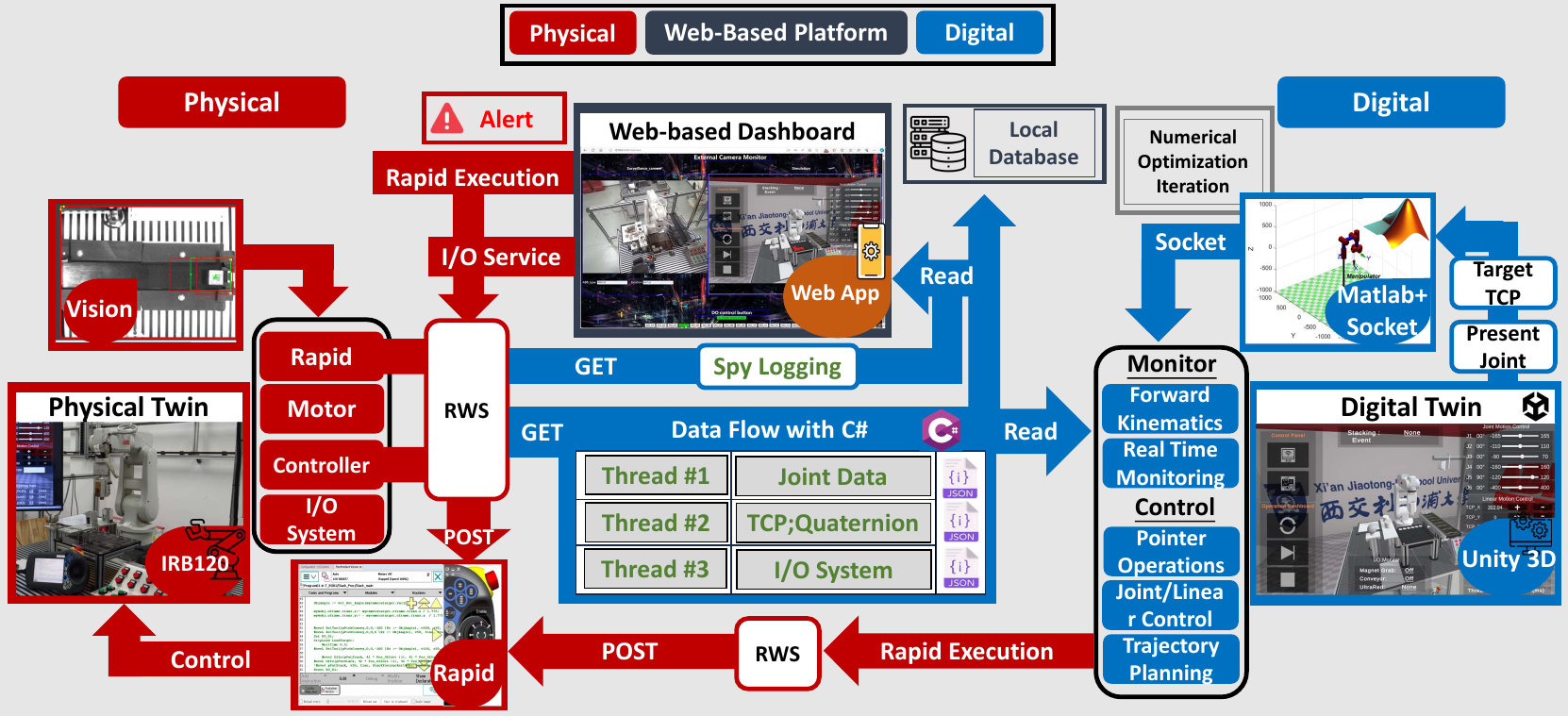}
    \caption{The Framework of the Proposed Digital Twin System for the Stacking Robot Workstation}
    \label{Framework}
\end{figure*}

As shown in Fig. \ref{GUI}, the robot manufacturing station is designed initially for grasping, handling and stacking of objects, which can, however, easily be reconfigured for other manufacturing tasks, e.g. assembly, grinding/polishing, 3D printing, etc. This robot manufacturing station is also known as the physical twin as given in Fig. \ref{Framework}. The manufacturing station is equipped with a 6DOF robot arm (the ABB IRB120 industrial robot arm with the IRC5 compact controller \cite{abbmain}), a magnetic suction gripper for handling metallic blocks, a 2D camera (COGNEX 821-0084-5R \cite{cognexStandardComponents}) for objection recognition and offset calculation. The other accessories include a conveyor belt driven by a DC motor and a palletized platform for deposition of objects being handled. This paper consider metallic workpieces of 3 geometries (square, rectangle, circle) to be handled and stacked.

The workpiece is initially placed on the part on Location A of the conveyor belt, where it's detected by a camera capturing its features. Upon detection, the part moves to Location B, halting the conveyor via an infrared sensor. The robot arm then uses camera data to pick up and place the part on the platform, before returning to its starting position for the next cycle.

ABB Robot Studio \cite{robotstudio}, Teach FlexPendant, as well as ABB Integrated Vision system have been used for program development. The robot motion program is based on RAPID programming language \cite{2023rapid}. The Rapid system is embedded in FlexPendant, supervising various tasks. Specifically, Rapid has the capability to select different T\_ROB (a task or program defining a specific sequence for the robot), adjust the PP (pointer), override the Digital Input (DI) and Digital Output (DO) systems (I/O System), etc. Furthermore, I/O system is used for virtual environment to read abstract events. For example, DO\_3, DO\_4, and DO\_5 that configured to three indicator lights, are used to connect the recognition result, enabling Unity3D monitor and control the different abstract manufacturing events.

802.11n, which is a wireless networking standard introduced by IEEE \cite{2024ieee}, has been used for data transmission. Based on the standard, we can develop data transmission mechanism between the physical platform and the corresponding virtual simulator with a high communication efficiency. Meanwhile, we have applied Multiple Input Multiple Output (MIMO) technology, which employs multiple antennas to transmit and receive data simultaneously. These features can well fit our expectations as they can enable efficient communication inside the DT framework, which is the main novelty of this paper.

\subsection{The Digital Twin Platform (Unity 3D) of the Robot Manufacturing Station}

\begin{figure*}
\begin{center}
    \includegraphics[width= 16cm]{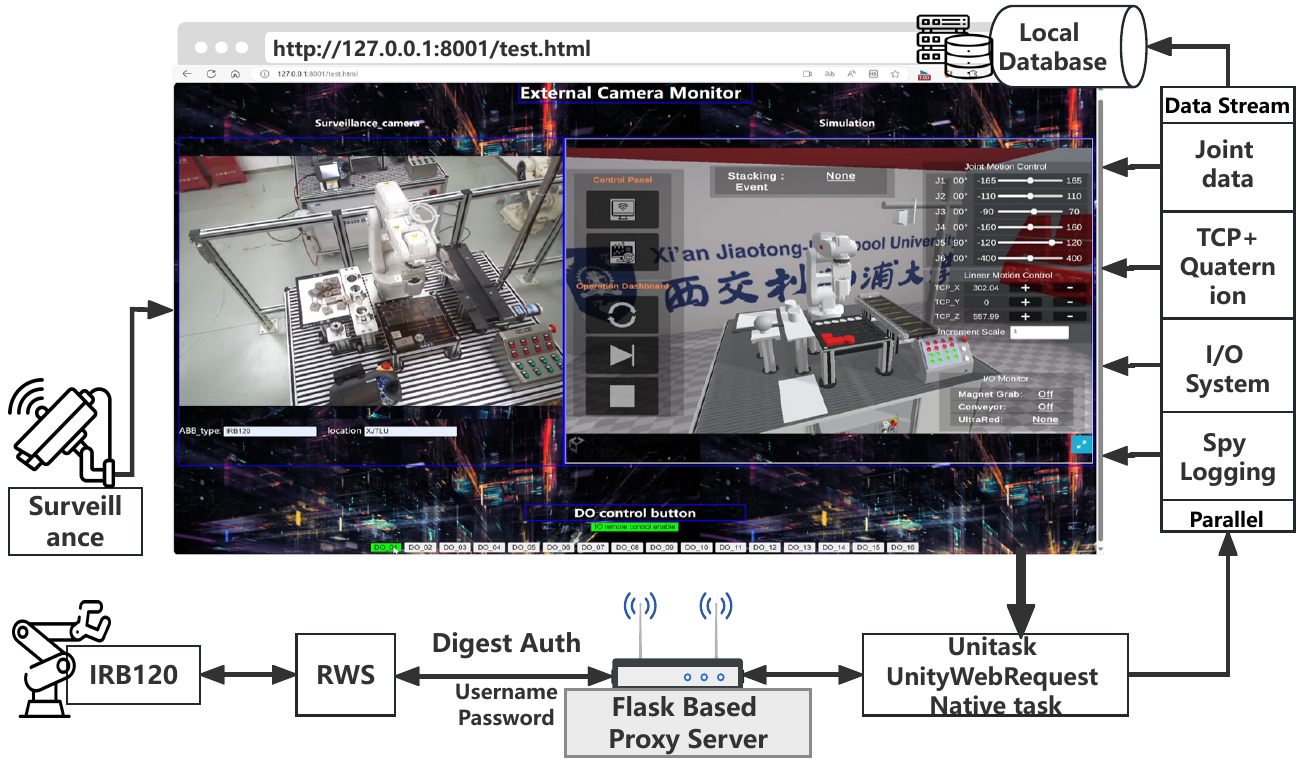}
\caption{Web-based Platform Framework Design and Working Screenshot} 
\label{Fig.9 WEB}
\end{center}
\end{figure*}

As shown in Fig. \ref{GUI}, the DT platform is developed by constructing a virtual environment in Unity 3D with all necessary components available in the physical robot station. The 3D models of main components are developed using Blender exported as FBX format. Apart from the virtual model of the station, we have developed a simple virtual laboratory space and GUIs that enable real-time control of the workstation and display necessary parameters during operation.

However, to ensure the precise robot model kinematics, we firstly obtain the IRB120 STEP model from \cite{abbmain}, then transform it to STL file to blender with absolute accuracy. Next, blender can add  parent's relation to each link's while export it to FBX format to Unity 3D. Therefore, we can implement model forward kinematics visualization. It relies on Transform.localEulerAngles class in Unity 3D \cite{unity3dUnityScripting}. It has the advantage that represent a three-dimensional rotation by performing three separate rotations around individual axes. Therefore, each transform is set to the Motor shaft Center at each link, while every link's parent is set to up level link. Finally, we assign the obtained Joint Data to each link's script that has Transform.localEulerAngles rotation. The model kinematics is achieved.

The data stream that includes three threads, i.e. (1) joint motion parameters; (2) Tool of Center Points (TCP) and Quaternion; (3) Data from the robot controller I/O system, is the core component of the Fig. \ref{Framework} framework, providing key parameters for the real-time simulation of robot workstation.

On the other hand, data that feedback to physical twin are (1) Pointer operation command, and (2) calculated joint motion parameters obtained updating command. The desired pointer operation command (1) is sent from Pointer Operation GUI in Unity 3D. Further, for the desired joint motion parameters (2), a framework which interacts via socket, utilizing Unity 3D, together with MATLAB, including our algorithms conducting real-time path planning also has been integrated into the DT platform. This framework's input is from Motion GUI defining multiple-joints motion or X, Y, Z linear motions before generating the desired joint motion parameters.

\subsection{Web-Based Platform}

With the advantage of RWS that based on Rest API, we propose a framework that uses WebGL to construct Web-based Platform. However, translating Unity project to JavaScript using WebGL can result in errors. Specifically, original C\# functions like networking and threading are not supported on the web. Furthermore, replacing threads with UniTask can cause back-end platform to crash as the back-end platform will be overwhelmed by flood of data. Additionally, replacing networking with UnityWebRequest is challenging due to the absence of function that creates digest authorization, a fundamental part of RWS system.

To address these issues, we have developed a Web-Based Platform, based on a specific framework, all of which are shown in Fig \ref{Fig.9 WEB}. This structure has implemented a proxy server using Flask by Python. This proxy server provides functions that are more easily transmit digest authorization to ABB through RWS, allowing requests from the web to Flask and ultimately reaching ABB.

With our developed Web-Based platform, users can not only leverage all functions available in Unity3D, but also use the following new functions. 
\begin{itemize}
    \item Users can adjust the DO system.
    \item A surveillance Camera Monitoring is integrated into Web-based Platform.
    \item Spy Logging is integrated with main data which enables users to observe production events in workstation.
\end{itemize}

\subsection{Communication and Data Transmission}

Both Unity 3D and Web-based platform request the data streams, as shown in Fig. \ref{Framework} \ref{Fig.9 WEB}. While the data stream plays a crucial role in virtual environment framework construction, they are all based on RWS using the HTTP protocol and a C\# based client \cite{robot}. This design results in high data communication efficiency and accuracy. All responses to read requests are in JSON format, GET Method, and digest authorization. There are four thread types constructing data stream in framework:

\begin{itemize}

    \item Thread \#1: GET real-time joints data via RWS's "Get Motion System action" \cite{robot}, utilized for forward kinematic simulation, trajectory planning, etc.
    
    \item Thread \#2: GET Tool of Center Point (TCP) and Quaternion data via RWS's "Get Motion System action" \cite{robot}, employed for user-friendly GUI (Joint Motion Control and Linear Motion Control).

    \item Thread \#3: GET I/O system data to create abstract event via RWS's "Get IO System resource" \cite{robot}. Then abstract events are employed to simulate various components in DT, such as conveyor, infrared sensor, and magnetic sucker, etc. These enable a comprehensive DT stacking virtual environment with real-time comprehensive monitoring and control.

    \item Thread \#4: GET spy Logging (Exclusive in Web-based Platform) via RWS's "Start RAPID Spy Logging" \cite{robot}. Spy logging is the events in the RAPID production panel, which is added to Web-based platform, to enable events monitoring.

\end{itemize}

On the other hand, the control functions within the virtual environment shift the RWS communication protocol to the POST method. The RAPID teach FlexPendant plays a crucial role because all remote control commands are based on it. There are four available control functions:

\begin{itemize}
    \item Pointer Operation (Fig. \ref{GUI} 1st section), achieved via RWS's "Operations on RAPID execution" \cite{robot}, which entails: (1) Resetting RAPID program pointer to main; (2) Starting RAPID Execution; (3) Stopping RAPID Execution.
    
    \item Joint Motion Control (Fig. \ref{GUI} 4th section), achieved via RWS's "Update rapid variable current value" \cite{robot}. To cooperate with the updated joint value, the RAPID program must be amended to a TOB equipped with RAPID Motion Control Command, "MoveABSJ", by modifying PP.

    \item Linear Motion Control (Trajectory Planning) (Fig. \ref{GUI} 5th section): Enable users to adjust the TCP via Linear Motion in X,Y,Z direction by a proposed framework which will be illustrated later.

    \item DO System Remote Control (Exclusive in Fig. \ref{Fig.9 WEB} Web-based Platform): This is achieved via RWS "Update IO Signal Value feature" \cite{robot} to adjust the DO terminal value.
    
\end{itemize}

\section{Motion Control Algorithm}

\begin{figure}[]
    \centering
    \includegraphics[width = \linewidth]{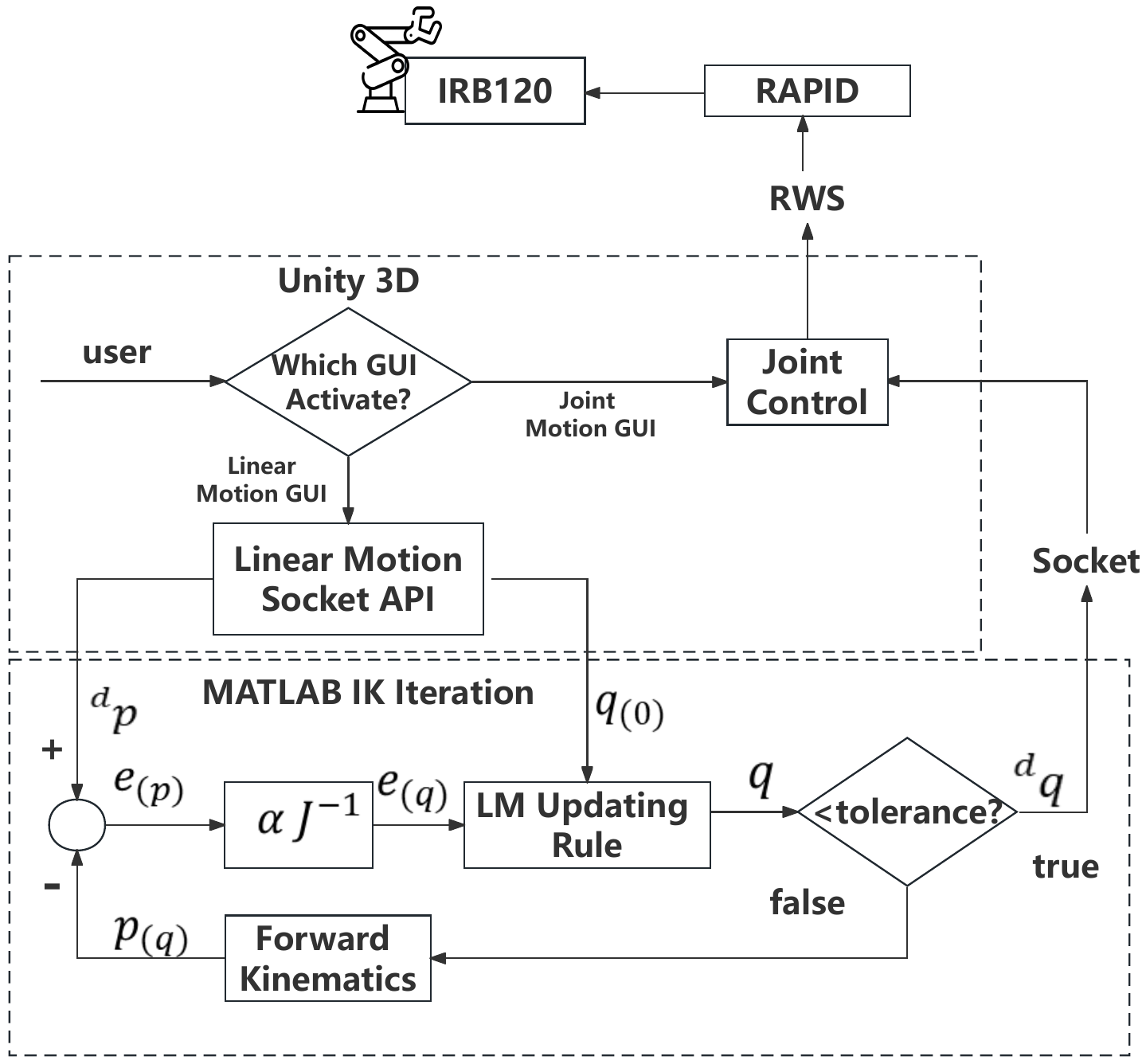}
    \caption{The Framework and Algorithm of Motion Control across Unity 3D, MATLAB, Robot Arm IRB120}
    \label{Numerical solution figure}
\end{figure}

Our proposed motion control under DT framework is shown in Fig. \ref{Numerical solution figure}. It is visualized as GUI in Unity 3D that have two ways to adjust the position of the robot arm, including (1) Joint Control and (2) Linear Control. Joint Motion control can be easily implemented. After knowing the expected joint space value of target point from Unity 3D, DT can send the data to the RAPID FlexPendant by updating the data in specified T\_ROB by RWS. Then robot arm execute "MoveABSJ" motion control command to obtain the target position directly. However, because the "MoveABSJ" command can only receive joint space value, the Linear Motion cannot be directly implemented, which means Inverse Kinematics (IK) is compulsory to solve the available joint space value to the robot controller. We therefore developed and integrated into our proposed DT Kinematics method, which is based on Levenberg Marquardt (LM) numerical Inverse Kinematics method.


\subsection{Levenberg–Marquardt Method of Inverse Kinematics}

MATLAB is set as TCP/IP server to receive data. Upon Linear Motion Unity GUI is activated, Unity 3D transmits the desired Tool of Center Point (TCP) to be reached $^d \boldsymbol{p}$, together with joint configuration vector value at the same moment to MATLAB, serving as initial guess value $\boldsymbol{q}_{(0)}$. The robot arm parameters for ABB IRB120 like standard D-H table and Jacobin matrix can be easily found in \cite{xia2022abbirb120}. Throughout the IK iteration process, algebraic constraint on the joint configuration vector is denoted as $ \boldsymbol{q}=\left[q_1 q_2 \ldots q_6\right]^{\mathrm{T}} \in \mathbb{R}^6$.

Next, the joint space residual of the $i$-th position constraint is defined as ${e}_i$.

\begin{equation}
\boldsymbol{e}_i(\boldsymbol{q}) \equiv  ^d\boldsymbol{p}_i-\boldsymbol{p}_i(\boldsymbol{q})  \text { (position constraint) } \\ 
\end{equation}

The residual vector $\boldsymbol{e}(\boldsymbol{q}) $ comprises six elements:

\begin{equation}
\boldsymbol{e}(\boldsymbol{q}) \equiv\left[\begin{array}{llll}
\boldsymbol{e}_1^{\mathrm{T}}(\boldsymbol{q}) & \boldsymbol{e}_2^{\mathrm{T}}(\boldsymbol{q}) & \ldots & \boldsymbol{e}_6^{\mathrm{T}}(\boldsymbol{q})
\end{array}\right]^{\mathrm{T}}
\end{equation}

Introducing Jacobin matrix $\boldsymbol{J}_k$, the error of joint space value $\boldsymbol{e}(\boldsymbol{q})$ yields:

\begin{equation}
\boldsymbol{e}(\boldsymbol{q}) = \boldsymbol{\alpha} \boldsymbol{J}_k^{-1}  \boldsymbol{e}(\boldsymbol{p}) 
\end{equation}

where $\boldsymbol{e}_k(\boldsymbol{p}) = ^{d}{\boldsymbol{p}}-\boldsymbol{p}_k(q)$. As illustrated in the feedback iteration in Fig. \ref{Numerical solution figure}, the desired TCP to reach is $^d \boldsymbol{p}$, and the desired Joint space value to reach is $ ^{d}\boldsymbol{q}$. For conventional numerical IK solution, the updating rule in each iteration is:
 
 \begin{equation}
\boldsymbol{q}_{k+1}=\boldsymbol{q}_k + \boldsymbol{\alpha}  \boldsymbol{J}_k^{-1} \boldsymbol{e}_k + \boldsymbol{q}_{(0)}
\end{equation}

where $k$ is the iteration cycle, $\boldsymbol{e}_k  \equiv \boldsymbol{e}({q_k})$, $\boldsymbol{q}_{(0)}$ is the initial Joint configuration from Unity 3D, and $\boldsymbol{\alpha}$ is the learning rate, accelerating convergence velocity of iteration. The initial guess, $\boldsymbol{q}_{(0)}$, is very essential at real IK calculation since it ensures the convergence and accuracy of the solution.

\begin{figure*}
    \centering
    \includegraphics[width = \linewidth]{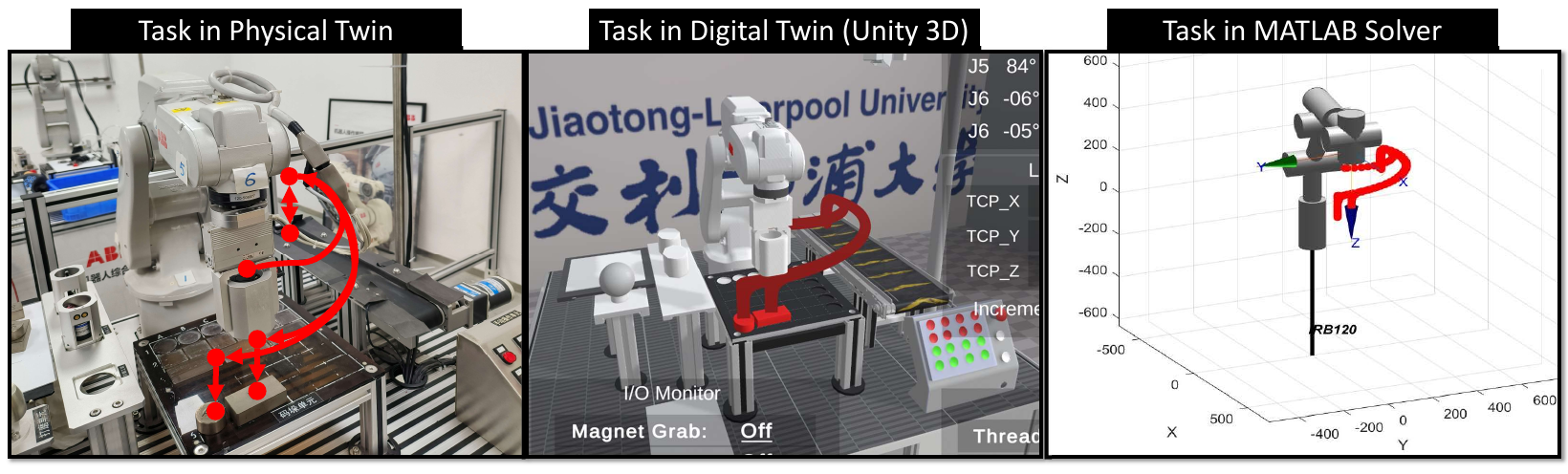}
    \caption{The Trajectory Mapping of the Robot Arm's Tool of Center Point across Physical Workstation, Digital Twin, and MATLAB Solver.}

    \label{Result_trajectory_mapping}
\end{figure*}

However, employing the conventional updating rate rule in the IK iteration process often leads to convergence towards local minimum values, and encountering singularities becomes inevitable with unsolvable original equations. Hence, the LM method is adopted, offering numerous advantages including enhanced convergence properties, robustness against ill-conditioned problems, and increased stability near singular points. This method also requires fewer iterations to achieve a solution. \cite{5784347}. The new updating method is represented to be equations (5) (6).

\begin{equation}
\boldsymbol{q}_{k+1} =\boldsymbol{q}_k+ \boldsymbol{\alpha} \boldsymbol{H}_k^{-1} \boldsymbol{g}_k + \boldsymbol{q}_{(0)}
\end{equation}

\begin{equation}
    \boldsymbol{H}_k  \equiv \boldsymbol{J}_k^{\mathrm{T}} \boldsymbol{W}_E \boldsymbol{J}_k+\boldsymbol{W}_N
\end{equation}

where $\boldsymbol{W}_N=\operatorname{diag}\left\{w_{N, i}\right\}\left(w_{N, i}>0\right.$ for $\left.\forall i=1 \sim n\right)$ represents the damping factor, and $\boldsymbol{g}_k \equiv \boldsymbol{J}_k^{\mathrm{T}} \boldsymbol{W}_E \boldsymbol{e}_k$. 

LM method updating rule ensures a decrease in the error function, as the matrix $\boldsymbol{H}_k$ is always both regular and positive definite. Consequently, the increment term of equation consistently directs towards the descent direction. Furthermore, the updating rule ensure that the mixed minimization problem of equation (7) could be solved in each step of iterations \cite{minimization}. 

\begin{equation}
\frac{1}{2} \boldsymbol{r}_k^{\mathrm{T}} \boldsymbol{W}_E \boldsymbol{r}_k+\frac{1}{2} \boldsymbol{e}{(\boldsymbol{q}_k^{\mathrm{T}})} \boldsymbol{W}_N  \boldsymbol{e}{(\boldsymbol{q}_k)} \rightarrow \min .
\end{equation}

where $ \boldsymbol{e}{(\boldsymbol{q}_k)} \equiv \boldsymbol{q}_{k+1} -\boldsymbol{q}_{k} $, and $ \boldsymbol{r}_k \equiv 
 \boldsymbol{e}_k - \boldsymbol{J}_k \boldsymbol{e}{(\boldsymbol{q}_k)} $. As a result, $q_k$ can converge to a certain value even in redundant cases, minimizing deviation. The LM method for minimization is preferred over the conventional method due to its superior numerical robustness in handling unsolvable cases prone to convergence towards singular points.

At the end of each iteration process, MATLAB calculates and judges whether the error between the TCP and the desired pose $ ( \boldsymbol{p}{(q)} - ^d\boldsymbol{p} )$ falls within the specified tolerance. If this is satisfied, joint space value to desired TCP value $^d\boldsymbol{q}$ is sent back to Unity 3D by establishing another TCP/IP server channel. Otherwise, iteration continues until the error falls below the specified tolerance. Finally, the expected joint space value $^d\boldsymbol{q}$ to desired TCP is transferred to Joint Control function in Fig. \ref{Numerical solution figure} mentioned earlier.

\section{Results and Analysis}

To validate the data interaction in proposed DT framework, a trajectory mapping of robot arm's TCP is conducted across real workstation, Digital Twin (Unity 3D), and MATLAB solver, as shown in Fig. \ref{Result_trajectory_mapping}, where the robot arm pauses at an intermediate position after stacking two workpieces. Notably, the trajectory lines exhibit high similarity. This finding proves the accuracy of the data flow design within the DT framework. Furthermore, it verifies the accuracy of the forward kinematics in Unity 3D and MATLAB solver. Hence, it also verifies the Inverse kinematics with real-time trajectory planning control solver in MATLAB.

\begin{figure}
    \centering
    \includegraphics[width = 8cm]{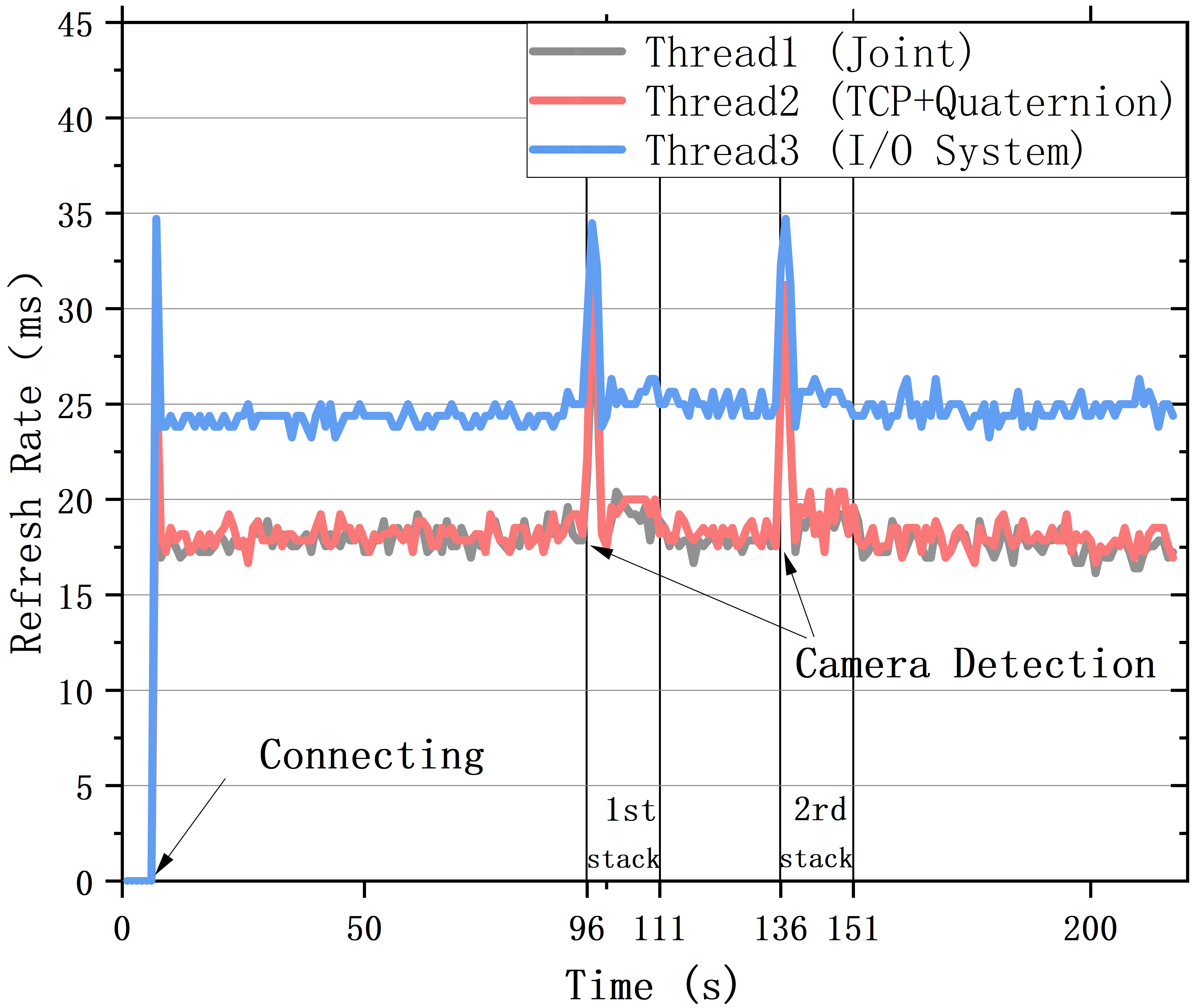}
    \caption{The Refresh Rate of Three Main Threads in Data Flow}
    \label{refresh rate}
\end{figure}

Fig. \ref{refresh rate} illustrates the real-time communication efficiency with refresh rate of these three threads during two stacking events (96-111s 1st stacking, 136-151s 2rd stacking). The refresh rate is defined as the reciprocal of the number of RWS GET requests per second by a counter in Unity 3D. Remarkably, even in the absence of PLC support at the workstation, thread 1 and thread 2 mainly maintain an impressive refresh rate of approximately $17 ms$, while thread 3 mainly maintains around $25 ms$ since high JSON file volume in I/O system communication.

\begin{table}[]
\caption{Linear Operation Contrast Experiment Result}
\label{Linear Control}
\resizebox{\linewidth}{!}{%
\begin{tabular}{c|ccc}
Control Method & Energy &  Accuracy  & 
 Reachability \\ \hline
RAPID Teach Pendant                       & $45.82J$                & 100\% &   100\%            \\
Digital Twin                 & $45.79J$                  & 100\% &  100\%           
\end{tabular}
}
\end{table}


However, the refresh rate tends to be higher upon initial connection. Because the thread counter not covering all GET requests at the program's outset. Furthermore, the efficiency of threads decreases notably when the ABB 2D vision camera initiates recognition of the workpiece type. And the three threads experiences the most significant delay when the camera calculates the offset position of the workpiece. This occurrence underscores that the device is so modest that even a simple 2D camera can exert a significant drain on system resources. However, these findings underscore the novelty of the data flow design, which ensures high communication efficiency even on a such modest device. Such high efficiency communication ensures the real-time synchronization of the model, and reliability of real-time Joint and Linear Control during continuous remote monitoring.

To further test the Motion Control functions in DT framework, Tab. \ref{Linear Control} conducts a contrast test. Users are prompted to adjust the TCP from the home position by $100mm$ along the X-axis through linear operations using both the ABB RAPID teach FlexPendant and the proposed linear control function. Energy consumption and X,Y,Z-axis values are tested using Robot Studio \cite{robotstudio}, capable of monitoring all parameters in the robot. Additionally, the accuracy and reachability of global operations for both control methods were found to be 100\%. The energy consumption are identical since the same dynamic method is employed. These results validate the successful implementation of the user-friendly Motion control GUI in DT system.

\section{Conclusion}

This paper have proposed a simple and novel approach of developing a digital twin framework which is generally applicable to industrial robotic systems in intelligent manufacturing. Our study considers an industrial robot workstation initially designed for pick-and-place, while the developed DT system includes the corresponding hardware robot station and a virtual model mapping developed in Unity 3D for real-time simulation. Dual-stream communication and data transfer is available in between the hardware robot station and the virtual model, which is set up by integrating RWS, while excluding any other 3rd party tools such as ROS or costly device PLC, which are considered unnecessary, thus simplicity is achieved with our approach.

This simple framework achieve very effective real-time communication in between the hardware robot station and the virtual environment with very quick refreshing rate of $17ms$, which is adequate for most robotics applications in manufacturing. Further, with our proposed framework, real-time control path planning of the robot gripper can be conducted via DT platform. We have developed a user-friendly GUI (in Unity 3D and Web) where we can set inputs for robot joints’ control, while the real-time trajectory planning is conducted by integrating a solver of the LM method Inverse Kinematics, implemented in MATLAB which is the most popular engineering software. To validate the reliability of rea-time control functionality with our developed DT, we have conducted the trajectory mapping while a contrast experiment between ABB Teach pendant and DT which reveals a 100\% accuracy and reachability of global control.

\bibliographystyle{IEEEtran}
\bibliography{digital_twin}

\begin{thebibliography}{10}
\providecommand{\url}[1]{#1}
\csname url@samestyle\endcsname
\providecommand{\newblock}{\relax}
\providecommand{\bibinfo}[2]{#2}
\providecommand{\BIBentrySTDinterwordspacing}{\spaceskip=0pt\relax}
\providecommand{\BIBentryALTinterwordstretchfactor}{4}
\providecommand{\BIBentryALTinterwordspacing}{\spaceskip=\fontdimen2\font plus
\BIBentryALTinterwordstretchfactor\fontdimen3\font minus \fontdimen4\font\relax}
\providecommand{\BIBforeignlanguage}[2]{{%
\expandafter\ifx\csname l@#1\endcsname\relax
\typeout{** WARNING: IEEEtran.bst: No hyphenation pattern has been}%
\typeout{** loaded for the language `#1'. Using the pattern for}%
\typeout{** the default language instead.}%
\else
\language=\csname l@#1\endcsname
\fi
#2}}
\providecommand{\BIBdecl}{\relax}
\BIBdecl

\bibitem{cimino2019review}
C.~Cimino, E.~Negri, and L.~Fumagalli, ``Review of digital twin applications in manufacturing,'' \emph{Computers in Industry}, vol. 113, p. 103130, 2019.

\bibitem{lu2020digitalb}
\BIBentryALTinterwordspacing
Y.~Lu, C.~Liu, K.~I.-K. Wang, H.~Huang, and X.~Xu, ``Digital twin-driven smart manufacturing: Connotation, reference model, applications and research issues,'' \emph{Robotics and Computer-Integrated Manufacturing}, vol.~61, p. 101837, 2020. [Online]. Available: \url{https://www.sciencedirect.com/science/article/pii/S0736584519302480}
\BIBentrySTDinterwordspacing

\bibitem{fuller2020digital}
A.~Fuller, Z.~Fan, C.~Day, and C.~Barlow, ``Digital twin: Enabling technologies, challenges and open research,'' \emph{IEEE Access}, vol.~8, pp. 108\,952--108\,971, 2020.

\bibitem{8477101}
F.~Tao, H.~Zhang, A.~Liu, and A.~Y.~C. Nee, ``Digital twin in industry: State-of-the-art,'' \emph{IEEE Transactions on Industrial Informatics}, vol.~15, no.~4, pp. 2405--2415, 2019.

\bibitem{10260401}
J.~Zhong, Z.~Wang, X.~Bao, X.~Zhou, D.~Wu, and Y.~Zheng, ``A digital twin monitoring system for belt conveyor,'' in \emph{2023 IEEE 19th International Conference on Automation Science and Engineering (CASE)}, 2023, pp. 1--7.

\bibitem{9926434}
Z.~Li, W.~Xu, J.~Liu, J.~Cui, and Y.~Hu, ``Digital twin-based virtual reconfiguration method for mixed-model robotic assembly line,'' in \emph{2022 IEEE 18th International Conference on Automation Science and Engineering (CASE)}, 2022, pp. 228--234.

\bibitem{martinez-gutierrez2021digital}
\BIBentryALTinterwordspacing
A.~Martínez-Gutiérrez, J.~Díez-González, R.~Ferrero-Guillén, P.~Verde, R.~Álvarez, and H.~Perez, ``Digital twin for automatic transportation in industry 4.0,'' \emph{Sensors}, vol.~21, no.~10, 2021. [Online]. Available: \url{https://www.mdpi.com/1424-8220/21/10/3344}
\BIBentrySTDinterwordspacing

\bibitem{s23010017}
\BIBentryALTinterwordspacing
R.~Chancharoen, K.~Chaiprabha, L.~Wuttisittikulkij, W.~Asdornwised, M.~Saadi, and G.~Phanomchoeng, ``Digital twin for a collaborative painting robot,'' \emph{Sensors}, vol.~23, no.~1, 2023. [Online]. Available: \url{https://www.mdpi.com/1424-8220/23/1/17}
\BIBentrySTDinterwordspacing

\bibitem{OPCUalowrefreshrate}
\BIBentryALTinterwordspacing
C.~Yang, X.~Tu, J.~Autiosalo, R.~Ala-Laurinaho, J.~Mattila, P.~Salminen, and K.~Tammi, ``Extended reality application framework for a digital-twin-based smart crane,'' \emph{Applied Sciences}, vol.~12, no.~12, 2022. [Online]. Available: \url{https://www.mdpi.com/2076-3417/12/12/6030}
\BIBentrySTDinterwordspacing

\bibitem{li2021automatic}
Y.~Li, K.~Wang, Y.~Zhai, and Q.~Zhang, ``Automatic {{Stacking System}} based on {{ABB Robots}} and {{Digital Twin Monitoring}},'' in \emph{Computing in AI, Internet of Things(IoT) and Computer Engineering Technology 2021}, 2021.

\bibitem{zhu2022robot}
\BIBentryALTinterwordspacing
Q.~Zhu, T.~Zhou, P.~Xia, and J.~Du, ``Robot planning for active collision avoidance in modular construction: Pipe skids example,'' \emph{Journal of Construction Engineering and Management}, vol. 148, no.~10, p. 04022114, 2022. [Online]. Available: \url{https://ascelibrary.org/doi/abs/10.1061/%28ASCE%29CO.1943-7862.0002374}
\BIBentrySTDinterwordspacing

\bibitem{wenna2022digital}
\BIBentryALTinterwordspacing
W.~Wenna, D.~Weili, H.~Changchun, Z.~Heng, F.~Haibing, and Y.~Yao, ``A digital twin for 3d path planning of large-span curved-arm gantry robot,'' \emph{Robotics and Computer-Integrated Manufacturing}, vol.~76, p. 102330, 2022. [Online]. Available: \url{https://www.sciencedirect.com/science/article/pii/S0736584522000199}
\BIBentrySTDinterwordspacing

\bibitem{robot}
\BIBentryALTinterwordspacing
Robot web services. [Online]. Available: \url{https://developercenter.robotstudio.com/api/rwsApi/index.html}
\BIBentrySTDinterwordspacing

\bibitem{2023opc}
``{O}{P}{C} {U}nified {A}rchitecture - {W}ikipedia --- en.wikipedia.org,'' \url{https://en.wikipedia.org/w/index.php?title=OPC_Unified_Architecture&oldid=1181851786}, [Accessed 20-03-2024].

\bibitem{PCSDK}
``{PC SDK},'' \url{https://developercenter.robotstudio.com/api/pcsdk/}, [Accessed 24-03-2024].

\bibitem{abbmain}
``{A}{B}{B},'' \url{https://global.abb/group/en}, [Accessed 09-06-2024].

\bibitem{cognexStandardComponents}
``{c}ognex-{s}tandard-{c}omponent,'' \url{https://support.cognex.com/}, [Accessed 23-03-2024].

\bibitem{robotstudio}
\BIBentryALTinterwordspacing
{{RobotStudio Suite}}. Robotics. [Online]. Available: \url{https://new.abb.com/products/robotics/robotstudio}
\BIBentrySTDinterwordspacing

\bibitem{2023rapid}
``{R}{A}{P}{I}{D} - {W}ikipedia --- en.wikipedia.org,'' \url{https://en.wikipedia.org/w/index.php?title=RAPID&oldid=1185609020}, [Accessed 20-03-2024].

\bibitem{2024ieee}
``{I}{E}{E}{E} 802.11n --- zh.wikipedia.org,'' \url{https://en.wikipedia.org/wiki/IEEE_802.11}, [Accessed 20-03-2024].

\bibitem{unity3dUnityScripting}
U.~Technologies, ``{U}nity api: {T}ransform.local{E}uler{A}ngles,'' \url{https://docs.unity3d.com/ScriptReference/Transform-localEulerAngles.html}, [Accessed 07-06-2024].

\bibitem{xia2022abbirb120}
G.~Xia, Z.~Xiao, and P.~Ji, ``{{ABB-IRB120 Robot Modeling}} and {{Simulation Based}} on {{MATLAB}},'' in \emph{2022 {{International Seminar}} on {{Computer Science}} and {{Engineering Technology}} ({{SCSET}})}, 2022, pp. 23--26.

\bibitem{5784347}
T.~Sugihara, ``Solvability-unconcerned inverse kinematics by the levenberg–marquardt method,'' \emph{IEEE Transactions on Robotics}, vol.~27, no.~5, pp. 984--991, 2011.

\bibitem{minimization}
\BIBentryALTinterwordspacing
Y.~Nakamura and H.~Hanafusa, ``{Inverse Kinematic Solutions With Singularity Robustness for Robot Manipulator Control},'' \emph{Journal of Dynamic Systems, Measurement, and Control}, vol. 108, no.~3, pp. 163--171, 09 1986. [Online]. Available: \url{https://doi.org/10.1115/1.3143764}
\BIBentrySTDinterwordspacing

\end{thebibliography}

\end{document}